\documentclass[journal]{IEEEtran}
\usepackage{cite}
\usepackage{amsmath,amssymb,amsfonts}
\usepackage{graphicx}
\usepackage{textcomp}
\usepackage{booktabs}
\usepackage{hyperref}
\usepackage{multirow}
\usepackage{float}
\usepackage{lscape}

\title{A Framework for Early Sepsis Prediction via Self-Supervised (JEPA) and Federated Representation Learning}

\author{
\IEEEauthorblockN{Umair bin Mansoor$^{1*}$\thanks{https://orcid.org/0000-0002-6986-3774}, Dr. Munaf Rashid$^{2\dagger}$\thanks{https://orcid.org/0000-0003-2063-4513}, Dr. Roomi Naqvi$^{3\dagger}$\thanks{ORCID placeholder}}\\
\IEEEauthorblockA{$^{1}$Electrical Engg. Department, DHA Suffa University, Karachi, Pakistan\\
$^{2}$Salim Habib University, Karachi, Sindh, Pakistan\\
$^{3}$Intel Foundry Services, Intel, Tempe, Arizona, United States\\
$^{*}$Corresponding author: umair.mansoor@dsu.edu.pk\\
$^{\dagger}$These authors contributed equally to this work.\\
Emails: sheikh.munaf@shu.edu.pk (M. Rashid), syed.naqvi@intel.com (R. Naqvi)}
}

\begin{document}

\maketitle

\begin{abstract}
Early sepsis prediction from electronic health records is challenged by irregular sampling, high missingness, and class imbalance. We systematically compare four modeling paradigms---self-supervised Joint Embedding Predictive Architecture (JEPA) via masked latent prediction, self-supervised VICReg (variance-invariance-covariance regularization) with two-view augmentation, semi-supervised fine-tuning of a VICReg-pretrained encoder, and supervised Temporal Convolutional Network (TCN)---alongside raw-feature baselines. All models share a common preprocessing pipeline of hourly binning with forward-fill imputation applied to 7 biomarkers selected via sparsity analysis from the MIMIC-III dataset. Our best model (JEPA + XGBoost + mean pooling) achieves AUPRC 0.636 at the time of onset (H0), approaching the SupMix benchmark (0.667) while using 83\% fewer biomarkers. The Tier 1 pipeline---VICReg pretraining followed by semi-supervised fine-tuning and XGBoost---achieves AUPRC 0.510 at H0, a 3.1$\times$ improvement over the raw-feature baseline (0.165) and a 7.6\% improvement over the end-to-end supervised TCN (0.474). Crucially, the fine-tuned VICReg encoder exhibits the most temporally persistent representations, degrading only 16.8\% from H0 to H10 compared to 47.5\% for supervised TCN and 65.3\% for JEPA, demonstrating that self-supervised pretraining with task-aware fine-tuning yields features that are both sharp near onset and robust across prediction horizons.
\end{abstract}

\section{Introduction}
Sepsis is a life-threatening organ dysfunction caused by a dysregulated host response to infection, affecting approximately 49 million people annually worldwide with mortality rates of 20--50\%~\cite{rudd2020}. Early detection and intervention are critical, as each hour of delayed antibiotic administration is associated with a 7.6\% increase in mortality~\cite{kumar2006}. Machine learning models operating on electronic health record (EHR) data offer a promising avenue for early sepsis prediction. However, EHR time series present several challenges: (1) irregular sampling intervals, (2) high rates of missingness (often exceeding 70\% for laboratory values), (3) varying clinical practices across institutions, and (4) class imbalance (sepsis prevalence is typically 5--15\% in ICU populations).

Prior work has explored various architectural choices for this task. Futoma et al.~\cite{futoma2017} proposed a multi-output Gaussian process recurrent neural network (MGP-RNN) achieving AUPRC $\sim$0.50--0.55 on Duke EHR data, using 34 physiological variables with end-to-end MGP imputation and an LSTM classifier. Moor et al.~\cite{moor2019} introduced the MGP-TCN on MIMIC-III data with 44 biomarkers, achieving AUPRC 0.35 at 7 hours before onset. Wanyan et al.~\cite{wanyan2023} achieved AUROC 0.883 and AP 0.667 on the PhysioNet 2019 dataset using sub-categorical supervised pretraining (SupMix) with XGBoost. However, all of these prior works used 34--44 biomarkers, many with extreme sparsity ($<$5\% observation rate). Desautels et al.~\cite{desautels2016} demonstrated that even a minimal set of vital signs and age can provide clinically useful sepsis predictions with their InSight system on MIMIC-III, supporting the premise that sparse but carefully selected inputs retain discriminative power.

In this work, we take a deliberately different approach: we restrict our analysis to only 7 biomarkers selected through a systematic sparsity analysis. This strategy ensures that each biomarker contributes meaningful signal rather than noise from imputation. We systematically compare four modeling paradigms applied to the same preprocessed data: (1) a self-supervised JEPA trained via masked latent prediction, (2) a self-supervised VICReg with two-view augmentation, (3) semi-supervised fine-tuning of the VICReg-pretrained encoder with a BCE classification head, (4) an end-to-end supervised TCN, and (5) raw feature baselines with conventional classifiers. We investigate the impact of pooling strategy (mean vs.\ last-timestep) and classifier choice (Logistic Regression vs.\ XGBoost). We evaluate all models across prediction horizons $H = \{0, 1, \dots, 10\}$ hours before sepsis onset.

\section{Related Work}

\subsection{Clinical Early Warning Scores}
Traditional clinical scoring systems such as NEWS, MEWS, SIRS, and qSOFA have been widely deployed~\cite{smith2013, singer2016}. These scores compare a small number of physiological variables to fixed thresholds and assign independent scores, ignoring complex temporal interactions. Futoma et al.~\cite{futoma2017} showed that NEWS produced 2.5 false alarms per true alarm at sensitivity 0.85, compared to 0.5 for their MGP-RNN.

\subsection{Deep Learning for Sepsis Detection}
Futoma et al.~\cite{futoma2017} proposed the MGP-RNN framework, which uses a Multi-task Gaussian Process (MGP) to interpolate irregularly sampled multivariate time series onto a regular grid, then feeds the imputed values into an LSTM classifier. The model is trained end-to-end using the Lanczos method for scalable GP sampling. On a cohort of 49,312 inpatient admissions with 34 physiological variables, 35 baseline covariates, and 8 medication classes, they achieved 19.4\% AUROC and 55.5\% AUPRC improvement over NEWS.

Moor et al.~\cite{moor2019} replaced the LSTM with a TCN, producing the MGP-TCN. On MIMIC-III data~\cite{johnson2016} with 44 biomarkers, they extracted Sepsis-3 labels at hourly resolution. After case-control matching (570 cases, 5,618 controls), MGP-TCN achieved AUPRC 0.35 at 7 hours before onset, while their DTW-KNN ensemble reached 0.40. Lee et al.~\cite{lee2024} later improved upon MGP-TCN by incorporating GRU-D imputation, achieving AUPRC 0.710 on a Korean ICU cohort. Separately, Apalak and Kiasaleh~\cite{apalak2024} demonstrated that TCNs can also leverage ECG-derived heart rate variability features for sepsis prediction, broadening the range of physiological signals amenable to temporal convolution-based approaches.

Wang and Yao~\cite{wang2022} proposed a Multi-Branching TCN (MB-TCN) that handles missing values via a learned mask and addresses class imbalance through multiple balanced branching outputs with a shared core TCN. On the PhysioNet 2019 Challenge dataset with 40 clinical variables, MB-TCN achieved AUROC 0.892 and AUPRC 0.527, outperforming LSTM-based baselines by 7\% in AUROC and 27\% in AUPRC. Our approach achieves higher AUPRC (0.636 vs.\ 0.527) despite using 83\% fewer biomarkers, though their higher AUROC (0.892 vs.\ 0.838) highlights the remaining value of dense laboratory features for discriminative calibration.

Separately, Dai et al.~\cite{dai2023} proposed PoEMS, a deep reinforcement learning framework that augments an LSTM-based sepsis predictor with a policy network that decides adaptively when to issue a warning. On the PhysioNet 2019 dataset, PoEMS achieved AUROC 0.911 and AUPRC 0.675, with a tunable parameter $\alpha$ controlling the earliness--accuracy trade-off at inference time. Unlike their RL-based adaptive stopping, our work systematically evaluates fixed prediction horizons (H0--H10) and investigates whether self-supervised pretraining can produce temporally persistent representations across all horizons simultaneously.

Wanyan et al.~\cite{wanyan2023} proposed SupMix, a sub-categorical supervised pretraining strategy. Using XGBoost on the learned representations, they achieved AUROC 0.883 and AP 0.667 on PhysioNet 2019.

D\"using and Cimiano~\cite{dusing2025} proposed a federated attention-enhanced LSTM with variable prediction horizons, demonstrating that collaborative training across institutions is particularly beneficial for long-horizon sepsis prediction.

Beyond retrospective studies, Boussina et al.~\cite{boussina2024} deployed the COMPOSER deep-learning model across two emergency departments at UC San Diego, achieving AUROC 0.938--0.945 for 4-hour-ahead sepsis prediction. In a before-and-after quasi-experimental study of 6,217 septic patients, deployment was associated with a 1.9\% absolute mortality reduction and a 5.0\% increase in sepsis bundle compliance. Subsequently, Shashikumar et al.~\cite{shashikumar2025} extended COMPOSER with an LLM-based pipeline (COMPOSER-LLM) that processes unstructured clinical notes, achieving sensitivity 72.1\% and PPV 52.9\% with only 0.0087 false alarms per patient-hour, demonstrating the complementary value of unstructured data for sepsis prediction.

\subsection{Self-Supervised Learning for Time Series}
Contrastive Predictive Coding (CPC)~\cite{oord2018} learns representations by predicting future latents. VICReg~\cite{bardes2022} avoids negative samples by combining invariance, variance, and covariance regularization on two augmented views of the same input. Tonekaboni et al.~\cite{tonekaboni2021} proposed temporal neighborhood coding for clinical time series. Yang et al.~\cite{yang2026} developed MSTCL, a multi-scale temporal-aware contrastive learning model using only six vital signs, achieving AUROC 88.34\% for online sepsis prediction. Chen et al.~\cite{chen2025} proposed SCMCLR, a cross-modal contrastive learning framework that transfers knowledge from laboratory tests to vital sign representations, enabling accurate prediction using only vital signs. Most recently, Wu et al.~\cite{wu2026visreg} showed that VICReg's covariance term captures only second-order statistics and proposed VISReg, which replaces it with a Sliced-Wasserstein sketching objective for full distributional alignment, yielding improved representation quality across diverse data regimes.

\section{Methodology}

\subsection{MIMIC-III Dataset and PhysioNet 2019 Challenge}

\subsubsection{The MIMIC-III Database}
The Multiparameter Intelligent Monitoring in Intensive Care (MIMIC-III) database~\cite{johnson2016} comprises de-identified clinical data from over 58,000 hospital admissions of 46,520 patients admitted to the Beth Israel Deaconess Medical Center (BIDMC) between 2001 and 2012. The database is organized into multiple relational tables: \texttt{ADMISSIONS} (demographic and admission details), \texttt{ICUSTAYS} (ICU episode start and end times), \texttt{CHARTEVENTS} (bedside vital sign measurements recorded approximately hourly), \texttt{LABEVENTS} (laboratory test results drawn at clinician-determined intervals), \texttt{PRESCRIPTIONS} (medication orders), \texttt{PROCEDURES\_ICD} (ICD-9 procedure codes), and \texttt{DIAGNOSES\_ICD} (ICD-9 diagnosis codes). MIMIC-III is a publicly available de-identified database hosted on PhysioNet (https://doi.org/10.13026/C2XW26). The original data collection was approved by the Institutional Review Board of the Massachusetts Institute of Technology (MIT IRB), and all patient records were de-identified in accordance with the Health Insurance Portability and Accountability Act (HIPAA) safe harbor provisions.

\subsubsection{The PhysioNet 2019 Challenge Dataset}
The PhysioNet/Computing in Cardiology Challenge 2019~\cite{reyna2020} processed a subset of MIMIC-III data (augmented with data from Emory University Hospital) into a standardized, hourly-resolution format for sepsis prediction. The preprocessing pipeline involved: extracting all ICU stays from MIMIC-III; querying \texttt{CHARTEVENTS} for vital signs (heart rate, blood pressure, temperature, respiratory rate, SpO2) and \texttt{LABEVENTS} for laboratory tests (WBC, creatinine, lactate, bilirubin, etc.); aligning all measurements to hour boundaries relative to ICU admission; and computing hourly aggregates. The final dataset contains 40,336 ICU stays with 34 physiological variables, patient demographics (age, gender), hospital admission timing, and a binary SepsisLabel per hour.

Sepsis is defined according to Sepsis-3 criteria~\cite{singer2016}: a suspected infection (blood culture draw and antibiotic administration within $\pm$24 hours) accompanied by a Sequential Organ Failure Assessment (SOFA)~\cite{vincent1996} score increase of at least 2 points within the window $[-48, +24]$ hours around the suspicion of infection. The SOFA score is computed from six components: respiratory (PaO2/FiO2 or SpO2/FiO2 ratio), coagulation (platelet count), hepatic (bilirubin), cardiovascular (mean arterial pressure and vasopressor requirement), neurological (Glasgow Coma Scale), and renal (creatinine and urine output). Each component contributes 0--4 points, and the baseline SOFA score is taken as the minimum in the 48 hours preceding the suspicion of infection.

\subsubsection{Per-Patient Temporal Window Construction}
For each patient encounter, the time series is anchored to the ICU admission time ($t = 0$). Measurements are available as irregularly timed observations $\{(t_{ij}, v_{ij})\}$ where $t_{ij}$ is the clock time of the $j$-th observation of variable $i$ and $v_{ij}$ is its value. The challenge dataset resamples these into hourly bins indexed by $h = 0, 1, \dots, H_{\max}$, where $h = \lfloor t \rfloor$ is the integer hour since admission. If multiple observations of the same variable fall within a single hourly bin, they are reduced to summary statistics (mean, minimum, maximum, etc.). SepsisLabel $y_h \in \{0, 1\}$ is assigned per hour, where $y_h = 1$ for all hours after sepsis onset. The training/validation split follows a chronological protocol: admissions are ordered by admission time, with the first 37,897 stays allocated to training and the remaining 2,439 to validation, ensuring no temporal leakage.

\begin{figure*}[t]
\centering
\includegraphics[width=0.85\textwidth]{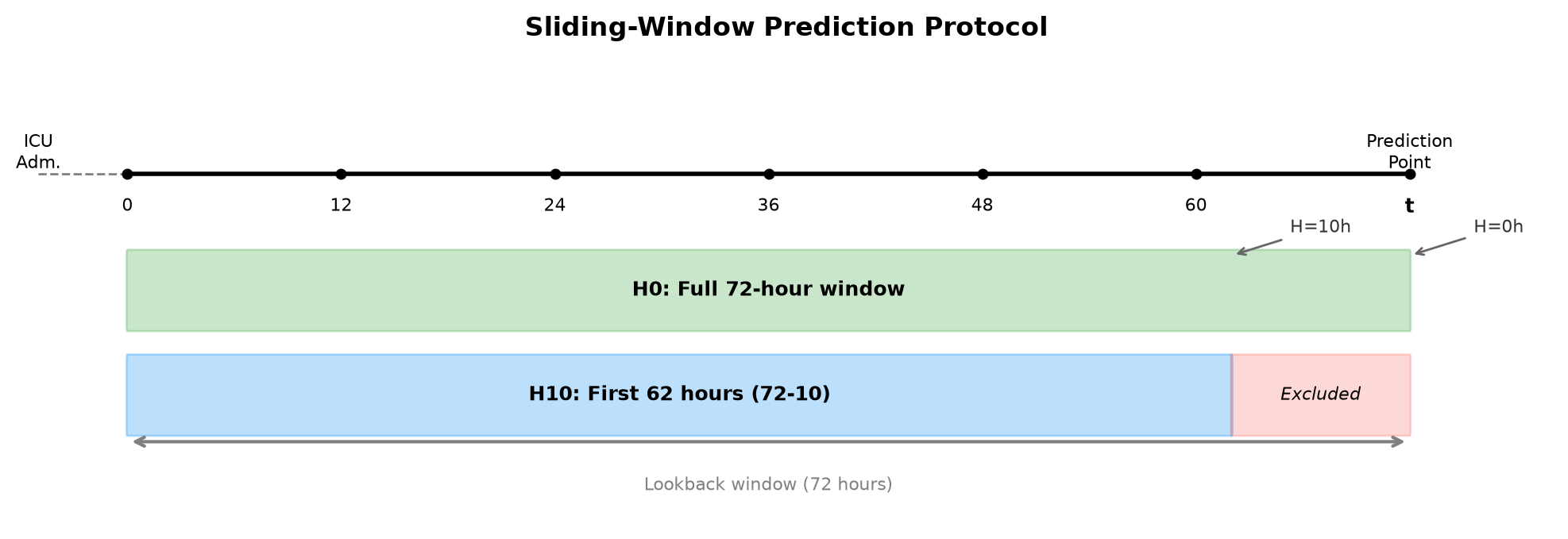}
\caption{Sliding-window prediction protocol. The 72-hour lookback window preceding the prediction time $t$ is progressively truncated as the horizon $H$ increases. At H0, all 72 hours are used; at H10, only the first 62 hours remain.}
\label{fig:horizon}
\end{figure*}

For the prediction task, we extract for each patient a 72-hour lookback window $\mathcal{W}_p = [t - 72, t)$ preceding the prediction time $t$. This window is divided into $T = 72$ hourly bins. At prediction horizon $H$, the input is restricted to the first $72 - H$ bins (i.e., data up to $H$ hours before the prediction point). This protocol, illustrated in Fig.~\ref{fig:horizon}, follows the evaluation methodology of Moor et al.~\cite{moor2019}.

\subsubsection{Biomarker Selection and Sparsity Analysis}
The PhysioNet 2019 dataset contains 34 physiological variables. We computed the fraction of hourly bins with at least one non-missing observation for each variable. Table~\ref{tab:sparsity} presents the results.

\begin{table*}[t]
\centering
\caption{Hourly observation rates for all 34 biomarkers. The 7 selected biomarkers (bold) have rates sufficient for reliable hourly binning. MAP (87.9\%) is included because its near-continuous availability complements SBP/DBP; WBC (6.4\%) is excluded due to sparsity.}
\label{tab:sparsity}
\small
\begin{tabular}{lrlrlr}
\toprule
\textbf{Biomarker} & \textbf{Rate} & \textbf{Biomarker} & \textbf{Rate} & \textbf{Biomarker} & \textbf{Rate} \\
\midrule
\textbf{HR}         & 90.6\% & BUN      & 7.0\% & Chloride      & 4.5\% \\
\textbf{SBP}        & 85.7\% & Hgb      & 7.4\% & Phosphate     & 4.2\% \\
\textbf{DBP}        & 69.0\% & pH       & 7.4\% & HCO3          & 4.4\% \\
\textbf{Temp}       & 34.0\% & Creatinine & 6.2\% & EtCO2        & 3.9\% \\
\textbf{RR / Resp}  & 85.3\% & Magnesium & 6.5\% & SaO2          & 3.9\% \\
\textbf{SpO2/O2Sat} & 87.8\% & WBC & 6.4\% & Lactate       & 3.1\% \\
\textbf{MAP}       & 87.9\% & Calcium  & 6.0\% & PTT           & 3.1\% \\
Glucose            & 16.5\% & Platelets & 5.9\% & BaseExcess    & 5.9\% \\
Potassium          & 9.3\%  & PaCO2    & 6.1\% & TroponinI     & 1.1\% \\
Hct                & 9.0\%  & AST      & 1.9\% & Fibrinogen    & 0.7\% \\
FiO2               & 8.5\%  & Alkalinephos & 1.8\% & Bilirubin dir. & 0.2\% \\
& & Bilirubin total & 1.7\% & & \\
\bottomrule
\end{tabular}
\end{table*}

Several observations guide our selection. First, the six vital signs (HR, SBP, DBP, Temp, RR, SpO2) are observed in 34--91\% of hourly bins because they are measured continuously or near-continuously at the bedside. Mean Arterial Pressure (MAP) is observed at 87.9\%, offering near-continuous blood pressure monitoring. Although MAP is a linear combination of SBP and DBP (MAP $\approx$ DBP + (SBP $-$ DBP)/3) and thus collinear, our 5-descriptor-per-biomarker representation (particularly the binary missingness flag) allows the TCN to learn the underlying linear relationship directly from data. Including MAP as a separate channel ensures the model receives the most complete blood-pressure signal possible even when SBP or DBP are transiently unrecorded.

Second, laboratory values have substantially lower observation rates because they depend on clinician-ordered blood draws. White Blood Cell Count (WBC, 6.4\%) is the most common sepsis-related laboratory test, but a 6.4\% hourly observation rate means that in 72-hour windows, a typical patient has WBC measurements in only 4--5 hours on average. The remaining 67 hours would contain the zero-descriptor placeholder $[0,0,0,0,0]$, making the WBC channel noise-dominated and degrading overall representation quality. No other laboratory marker (creatinine 6.2\%, lactate 3.1\%, bilirubin 1.7\%) exceeds 10\% observation, and all are therefore excluded.

Third, biomarkers with observation rates below 5\% (lactate, PTT, troponin, fibrinogen, bilirubin, etc.) are excluded because they would require $>$95\% synthetic imputation, introducing more noise than signal.

Thus, our final feature set comprises 7 biomarkers: HR, SBP, DBP, Temp, RR, SpO2, and MAP. These cover the cardiovascular (HR, SBP, DBP, MAP), respiratory (RR, SpO2), and metabolic (Temp) domains while avoiding extreme sparsity. The choice prioritizes near-continuous physiological monitoring over sparse laboratory signals, ensuring that each of the 42 hourly descriptors carries meaningful variance.

\subsection{Preprocessing: Absolute-Time Hourly Binning with Forward-Fill}
For each patient and each hourly bin, we compute 6 statistical descriptors for each of the 7 biomarkers, producing a 42-dimensional feature vector:
\begin{equation}
\mathbf{x}_h = [\mu_1, \sigma_1, n_1, l_1, m_1, t_1, \dots, \mu_7, \sigma_7, n_7, l_7, m_7, t_7] \in \mathbb{R}^{42}
\label{eq:features}
\end{equation}
where for biomarker $b$ at hour $h$: $\mu_b = \text{mean of observed values}$ (or last observed mean if forward-filled), $\sigma_b = \text{standard deviation}$ (0 if forward-filled), $n_b = \text{count of observations}$ (0 if forward-filled), $l_b = \text{last observed value}$, $m_b = \text{observed flag}$ (1 if at least one actual measurement in this hour, 0 if carried forward), and $t_b = \text{hours since last measurement}$ (0 if measured this hour). For hours with no new data, descriptors for each biomarker are forward-filled from the most recent hour containing an actual measurement. Biomarkers never observed during the window remain as zero vectors with $m_b = 0$ and $t_b = h$ (elapsed hours since window start). For the prediction task, we use a 72-hour lookback window.

\subsection{Temporal Convolutional Network (TCN) Encoder}
All models share a causal TCN encoder~\cite{bai2018}, which has been successfully applied to sepsis prediction in recent work~\cite{apalak2024, lee2024}. The dilated convolution is:
\begin{equation}
F(s) = (x \ast_d f)(s) = \sum_{i=0}^{k-1} f(i) \cdot x_{s-d\cdot i}
\label{eq:dilated}
\end{equation}
Each layer uses a residual block:
\begin{multline}
z^{(\ell+1)} = \mathrm{ReLU}\bigl(z^{(\ell)} + \\
\mathrm{Conv1D}(\mathrm{ReLU}(\mathrm{Conv1D}(z^{(\ell)})))\bigr)
\label{eq:residual}
\end{multline}
Our configuration: 5 residual blocks, channels [32, 64, 128, 256, 512], kernel size 3, dropout 0.2, dilation factors [1, 2, 4, 8, 16], latent dimension 128.

\subsection{Self-Supervised Pretraining via Masked Latent Prediction (JEPA)}
Our first self-supervised approach follows a masked autoencoder (MAE) paradigm applied to the latent space of the TCN encoder. For each 72-hour input sequence, we randomly select a contiguous block of $M$ hours as the target to be predicted:
\begin{equation}
M = \max(2, \lfloor \rho \cdot T \rfloor), \quad \rho = 0.4
\label{eq:mask_ratio}
\end{equation}
where $T = 72$ is the sequence length. The encoder processes the full sequence to produce latents $\mathbf{Z} = \{\mathbf{z}_1, \dots, \mathbf{z}_T\}$. A predictor network then reconstructs the latent representations of the masked block from the context (unmasked) latents. The predictor is a 2-layer MLP with residual connections, augmented with learnable positional embeddings and horizon embeddings:
\begin{equation}
\mathbf{z}^{\mathrm{pred}}_t = f_{\theta}\bigl(\mathbf{z}_t^{\mathrm{ctx}} + \mathbf{e}^{\mathrm{pos}}_t + \mathbf{e}^{\mathrm{hor}}_h \bigr) \quad \text{for masked } t
\label{eq:predictor}
\end{equation}
Masked positions are replaced with a learnable mask token~\cite{oord2018}. The pretraining objective combines prediction (MSE), variance, and covariance regularization on the predicted latents~\cite{bardes2022}:
\begin{multline}
\mathcal{L}_{\mathrm{JEPA}} = \frac{1}{n}\sum_{i}\|\mathbf{z}^{\mathrm{pred}}_i - \mathbf{z}^{\mathrm{target}}_i\|^2 \\
+ \lambda_v \frac{1}{d}\sum_{j} \max(0, \gamma - \mathrm{Std}(\mathbf{z}^{\mathrm{pred}}_j) + \epsilon) \\
+ \lambda_c \frac{1}{d}\sum_{i\neq j}[C(\mathbf{Z}^{\mathrm{pred}})]_{ij}^2
\label{eq:jepa}
\end{multline}
where $\lambda_v = 1.0$, $\lambda_c = 0.04$, $\gamma = 1.0$. Training: Adam optimizer, learning rate $5\times10^{-4}$, batch size 64, warmup 10 epochs, early stopping patience 30.

\subsection{Self-Supervised Pretraining via VICReg (Tier 1)}
Our second self-supervised approach uses VICReg~\cite{bardes2022}, which learns representations by maximizing agreement between two augmented views of the same input while preventing representational collapse through variance and covariance regularization. Unlike the masked autoencoder approach, VICReg does not require a predictor network or masking, making it computationally simpler.

For each input sequence $\mathbf{x}$ with mask $\mathbf{m}$, we generate two augmented views using a stochastic augmentation pipeline:
\begin{equation}
(\mathbf{x}_1, \mathbf{m}_1) = \mathcal{A}(\mathbf{x}, \mathbf{m}), \quad
(\mathbf{x}_2, \mathbf{m}_2) = \mathcal{A}(\mathbf{x}, \mathbf{m})
\label{eq:two_views}
\end{equation}
where $\mathcal{A}$ applies three augmentations: (1) \textit{temporal shift}: the input window is randomly shifted by up to $\pm 6$ hours with zero-padding at boundaries; (2) \textit{biomarker dropout}: each biomarker channel is independently dropped with probability 0.3 (replaced with zero descriptors); (3) \textit{Gaussian noise}: independent noise $\mathcal{N}(0, 0.02)$ is added to non-masked features.

Each view is encoded by the shared TCN encoder and pooled via masked mean pooling:
\begin{equation}
\mathbf{z}_k = \frac{\sum_{t=1}^{T} \mathbf{z}_{k,t} \cdot \mathbb{1}[\mathbf{m}_{k,t}]}{\sum_{t=1}^{T} \mathbb{1}[\mathbf{m}_{k,t}]}, \quad k \in \{1, 2\}
\label{eq:vicreg_pool}
\end{equation}

The VICReg objective has three components:
\begin{multline}
\mathcal{L}_{\mathrm{VICReg}} = \lambda_i \underbrace{\|\mathbf{z}_1 - \mathbf{z}_2\|^2}_{\text{invariance}} \\
+ \lambda_v \underbrace{\sum_{j=1}^{d} \max(0, \gamma - \mathrm{Std}(\mathbf{z}_j) + \epsilon)}_{\text{variance}} \\
+ \lambda_c \underbrace{\frac{1}{d} \sum_{i \neq j} [C(\mathbf{Z})]_{ij}^2}_{\text{covariance}}
\label{eq:vicreg}
\end{multline}
where invariance encourages the two views to map to the same representation, variance prevents dimensional collapse by pushing each dimension's standard deviation toward $\gamma=1.0$, and covariance decorrelates dimensions to prevent informational collapse. The covariance term enforces decorrelation but captures only second-order statistics, which recent work has identified as a limitation---VISReg~\cite{wu2026visreg} addresses this by replacing the covariance penalty with a Sliced-Wasserstein sketching objective that enforces full distributional shape, though we retain the standard VICReg objective for this study. We use $\lambda_i = 1.0$, $\lambda_v = 1.0$, $\lambda_c = 0.04$. Training: AdamW optimizer, learning rate $5\times10^{-4}$, batch size 64, weight decay $1\times10^{-5}$, cosine annealing schedule, early stopping patience 30.

\subsection{Semi-Supervised Fine-Tuning (Tier 1)}
After VICReg pretraining, we fine-tune the encoder for sepsis prediction by attaching a linear classification head:
\begin{equation}
\hat{y} = \sigma(\mathbf{W} \mathbf{z}_{\mathrm{pooled}} + b)
\label{eq:finetune_head}
\end{equation}
where $\mathbf{z}_{\mathrm{pooled}}$ is the masked mean-pooled latent. The classification head is a single linear layer (128$\to$1) with dropout (0.2). The entire model (encoder + head) is fine-tuned with binary cross-entropy loss:
\begin{equation}
\mathcal{L}_{\mathrm{BCE}} = -\frac{1}{N}\sum_{i=1}^{N}\bigl[y_i\log(\hat{y}_i) + (1-y_i)\log(1-\hat{y}_i)\bigr]
\label{eq:bce_ft}
\end{equation}
We use a lower learning rate ($1\times10^{-5}$) with AdamW, weight decay $1\times10^{-5}$, and early stopping with patience 15 to prevent catastrophic forgetting of the pretrained representations. The best checkpoint is selected based on validation loss.

After fine-tuning, the encoder is frozen and latents are extracted for downstream classification with XGBoost and Logistic Regression, following the same protocol as other models.

\subsection{Supervised TCN Training}
Binary cross-entropy loss:
\begin{equation}
\mathcal{L}_{\mathrm{BCE}} = -\frac{1}{N}\sum_{i=1}^{N}\bigl[y_i\log(\hat{y}_i) + (1-y_i)\log(1-\hat{y}_i)\bigr]
\label{eq:bce}
\end{equation}
The classification head applies global mean pooling followed by a linear layer (128$\to$1) with sigmoid. Training: Adam, learning rate $5\times10^{-4}$, batch size 64, warmup 10 epochs, early stopping patience 30.

\subsection{Classification Heads}
After representation learning, latents are pooled. Mean pooling: $\mathbf{z}_{\mathrm{mean}} = \frac{1}{T}\sum_{t=1}^{T} \mathbf{z}_t$. Last-timestep pooling uses $\mathbf{z}_T$ directly.

Logistic Regression with L2 regularization ($C = 1.0$):
\begin{equation}
P(y=1|\mathbf{x}) = \sigma(\mathbf{w}^T\mathbf{x} + b) = \frac{1}{1+e^{-(\mathbf{w}^T\mathbf{x}+b)}}
\label{eq:lr}
\end{equation}

XGBoost~\cite{chen2016}:
\begin{equation}
\mathcal{L}^{(t)} = \sum_{i=1}^{n}\ell\bigl(y_i, \hat{y}_i^{(t-1)} + f_t(\mathbf{x}_i)\bigr) + \gamma T + \frac{1}{2}\lambda\sum_{j=1}^{T}w_j^2
\label{eq:xgb}
\end{equation}
Configuration: 200 trees, max depth 4, learning rate 0.1, scale\_pos\_weight for imbalance.

\subsection{Evaluation Metrics}
AUPRC is the primary metric due to class imbalance~\cite{saito2015}:
\begin{equation}
\mathrm{AUPRC} = \int_{0}^{1} \mathrm{Precision}(r)\,dr
\label{eq:auprc}
\end{equation}
where $\mathrm{Precision} = TP/(TP+FP)$, $r = \mathrm{Recall} = TP/(TP+FN)$. We also report sensitivity at 90\% specificity.

\subsection{Federated Learning Setup}
We further investigate federated learning (FL) variants of the supervised TCN to simulate a decentralized clinical setting where patient data cannot be pooled due to privacy constraints. The training dataset is partitioned among 5 clients using Latent Dirichlet Allocation (LDA) with concentration parameter $\alpha = 0.5$, producing a heterogeneous label distribution: Client 0 (48.6\% positive, 346 samples), Client 1 (2.7\%, 915 samples), Client 2 (0.1\%, 1013 samples), Client 3 (0.5\%, 987 samples), and Client 4 (28.7\%, 470 samples). The two largest clients (1 and 2) contain the vast majority of negative-class samples, while Clients 0 and 4 carry most of the positive signal.

We implement and compare four FL aggregation strategies:
\begin{enumerate}
    \item \textbf{FedAvg}~\cite{mcmahan2017}: Server aggregates client model parameters via weighted averaging: $\mathbf{w}^{(t+1)} = \sum_{k=1}^K \frac{n_k}{N} \mathbf{w}_k^{(t)}$, where $n_k$ is the number of samples on client $k$ and $N$ is the total.
    \item \textbf{FedProx}~\cite{li2020fedprox}: Adds a proximal penalty $\frac{\mu}{2}\|\mathbf{w}_k - \mathbf{w}^{(t)}\|^2$ to the local loss, limiting per-client drift. We use $\mu = 0.01$.
    \item \textbf{FedDyn}~\cite{acar2021}: Maintains a per-client dynamic state $\mathbf{h}_k$ that corrects the local gradient: $\mathcal{L}_k^{\text{local}} = \mathcal{L}_k^{\text{BCE}} + \langle \mathbf{h}_k, \mathbf{w}_k \rangle + \frac{\alpha}{2} \|\mathbf{w}_k - \mathbf{w}^{(t)}\|^2$, with $\mathbf{h}_k \leftarrow \mathbf{h}_k - \alpha (\mathbf{w}^{(t)} - \mathbf{w}_k)$ after each round. We use $\alpha = 0.01$.
    \item \textbf{FedAvgM}~\cite{hsu2019}: Applies server-side momentum: $\mathbf{v}^{(t+1)} = \beta \mathbf{v}^{(t)} + \sum_k \frac{n_k}{N}(\mathbf{w}_k - \mathbf{w}^{(t)})$, $\mathbf{w}^{(t+1)} = \mathbf{w}^{(t)} + \eta \mathbf{v}^{(t+1)}$, with $\beta = 0.9$, $\eta = 0.3$.
\end{enumerate}
All FL variants use the same architecture, optimizer (AdamW, $\mathrm{lr}=5\times10^{-4}$), local epochs (1), and total communication rounds (50) for fair comparison. However, note that the FL experiments use a reduced architecture (encoder channels [16, 32], latent dimension 32) whereas the main results use the full architecture described in Section~III-C.

\section{Experimental Setup}
Experiments were conducted on CPU. JEPA pretraining: up to 100 epochs with early stopping (patience 30), learning rate $5\times10^{-4}$. VICReg pretraining: up to 100 epochs with early stopping (patience 30), AdamW optimizer, cosine annealing scheduler. Fine-tuning: up to 100 epochs with early stopping (patience 15), learning rate $1\times10^{-5}$, AdamW optimizer. Supervised TCN: up to 100 epochs with early stopping (patience 30), learning rate $5\times10^{-4}$. For horizon analysis, each model was trained once on available training data and evaluated at each $H \in \{0, \dots, 10\}$ by restricting the input window per the protocol of Moor et al.~\cite{moor2019}.

\section{Results and Discussion}

\subsection{Performance Summary}
Table~\ref{tab:results} presents AUROC and AUPRC for all experiments at horizons H0--H10 on the validation set (618 patients).

\subsection{Key Findings}
\textbf{Classifier Effect:} XGBoost dramatically outperforms Logistic Regression across all encoders. For JEPA, XGBoost improves AUPRC from 0.171 to 0.636 at H0 (+272\%).

For the fine-tuned VICReg encoder, XGBoost improves from 0.282 to 0.510 (+81\%). For supervised TCN, from 0.256 to 0.474 (+85\%). This confirms that tree-based classifiers are substantially more effective than linear models at predicting sepsis from temporal latent representations.

\textbf{Three-Tier Comparison:} The JEPA encoder achieves the highest H0 AUPRC (0.636), followed by the fine-tuned VICReg encoder (0.510, +7.6\% over the supervised TCN), then the supervised TCN (0.474), and finally the raw features (0.165). The Tier 1 pipeline---VICReg pretraining followed by semi-supervised fine-tuning---provides a meaningful intermediate between purely self-supervised (JEPA) and purely supervised (TCN) approaches, improving AUPRC by 7.6\% over end-to-end supervised training.

\textbf{Horizon Robustness:} The fine-tuned VICReg encoder exhibits the most stable performance across horizons. Its AUPRC drops only 16.8\% from H0 (0.510) to H10 (0.425), compared to 47.5\% for supervised TCN (0.474$\to$0.249), and 65.3\% for JEPA with XGBoost mean (0.636$\to$0.221). This suggests that the combination of self-supervised pretraining (which learns general physiological structure) and task-aware fine-tuning (which optimizes for discriminative features) produces representations that are both sharp near onset and persistent across horizons.

\subsection{Degradation Analysis Across Prediction Horizons}
We interpret the degradation hierarchy as follows. The self-supervised JEPA is optimized for masked latent reconstruction, which excels at capturing acute, localized temporal structure---the sharp physiological deterioration immediately preceding sepsis diagnosis. At H0, these acute signatures provide strong discriminative signal. However, as the horizon recedes, the acute signals fade and JEPA's representations, lacking task-aware pressure to learn subtle persistent patterns, degrade catastrophically.

Table~\ref{tab:degradation} quantifies the degradation asymmetry. The fine-tuned VICReg encoder loses only 16.8\% AUPRC from H0 to H10, making it the most horizon-robust learned representation. The supervised TCN loses 47.5\%, while the self-supervised JEPA loses 65.3\%.

\begin{table}[t]
\centering
\caption{AUPRC degradation from H0 to H10. The fine-tuned VICReg encoder is the most horizon-robust learned representation.}
\label{tab:degradation}
\scriptsize
\begin{tabular}{lccc}
\toprule
\textbf{Model} & \textbf{H0 AUPRC} & \textbf{H10 AUPRC} & \textbf{Drop} \\
\midrule
FT+XGB (mean) & 0.510 & 0.425 & \textbf{$-$16.8\%} \\
SupTCN+XGB (mean) & 0.474 & 0.249 & $-$47.5\% \\
JEPA+XGB (mean) & \textbf{0.636} & 0.221 & $-$65.3\% \\
JEPA+XGB (last) & 0.459 & 0.126 & $-$72.5\% \\
SupTCN+LR (mean) & 0.256 & 0.240 & $-$6.2\% \\
Raw+LR & 0.165 & 0.152 & $-$7.9\% \\
\bottomrule
\end{tabular}
\end{table}

The supervised TCN, trained end-to-end with BCE loss over the full 72-hour window, develops features under gradient pressure from each time step equally. This forces the encoder to identify patterns that remain discriminative regardless of their position relative to the prediction point, producing more temporally persistent features (47.5\% vs.\ 65.3\% drop).

The fine-tuned VICReg encoder achieves the best of both worlds: VICReg pretraining captures the general physiological structure through the two-view invariance objective, while the subsequent fine-tuning phase provides task-aware pressure that selects for discriminative patterns. The result is representations that are sharper at H0 than supervised TCN (0.510 vs.\ 0.474) and more robust across horizons than JEPA (16.8\% vs.\ 65.3\% drop). This suggests that self-supervised pretraining followed by lightweight task-aware fine-tuning is the most effective 

\begin{landscape}

\begin{table}[p]
\centering

\caption{Performance metrics across all experiments and horizons H0--H10. Classifiers: LR, XGBoost (XGB). Pooling: mean (m), last-timestep (l). Best result in bold.}
\label{tab:results}

\footnotesize                 
\setlength{\tabcolsep}{5pt}   
\renewcommand{\arraystretch}{1.25}

\begin{tabular}{lcccccccccccccccc}
\toprule

\multirow{2}{*}{H}
&
\multicolumn{2}{c}{Raw+LR}
&
\multicolumn{2}{c}{JEPA+LR}
&
\multicolumn{2}{c}{JEPA+XGB m}
&
\multicolumn{2}{c}{JEPA+XGB l}
&
\multicolumn{2}{c}{SupTCN+LR m}
&
\multicolumn{2}{c}{SupTCN+XGB m}
&
\multicolumn{2}{c}{FT+LR m}
&
\multicolumn{2}{c}{FT+XGB m}
\\

\cmidrule(lr){2-3}
\cmidrule(lr){4-5}
\cmidrule(lr){6-7}
\cmidrule(lr){8-9}
\cmidrule(lr){10-11}
\cmidrule(lr){12-13}
\cmidrule(lr){14-15}
\cmidrule(lr){16-17}

&
AUROC & AUPRC
&
AUROC & AUPRC
&
AUROC & AUPRC
&
AUROC & AUPRC
&
AUROC & AUPRC
&
AUROC & AUPRC
&
AUROC & AUPRC
&
AUROC & AUPRC
\\

\midrule

H0  & 0.713 & 0.165 & 0.709 & 0.171 & \textbf{0.838} & \textbf{0.636} & 0.798 & 0.459 & 0.759 & 0.256 & 0.801 & 0.474 & 0.800 & 0.282 & 0.850 & 0.510 \\
H1  & 0.717 & 0.167 & 0.690 & 0.155 & \textbf{0.876} & 0.611 & 0.637 & 0.117 & 0.744 & 0.245 & 0.790 & 0.361 & 0.782 & 0.213 & 0.839 & 0.469 \\
H2  & 0.719 & 0.171 & 0.685 & 0.149 & \textbf{0.858} & 0.562 & 0.680 & 0.134 & 0.739 & 0.243 & 0.771 & 0.311 & 0.780 & 0.211 & 0.839 & 0.470 \\
H3  & 0.716 & 0.168 & 0.684 & 0.147 & \textbf{0.827} & 0.500 & 0.621 & 0.128 & 0.737 & 0.242 & 0.768 & 0.288 & 0.778 & 0.208 & 0.834 & 0.466 \\
H4  & 0.716 & 0.166 & 0.674 & 0.146 & 0.819 & 0.429 & 0.614 & 0.109 & 0.733 & 0.241 & 0.759 & 0.261 & 0.775 & 0.204 & \textbf{0.831} & 0.464 \\
H5  & 0.712 & 0.164 & 0.675 & 0.141 & 0.817 & 0.406 & 0.607 & 0.109 & 0.726 & 0.238 & 0.730 & 0.240 & 0.774 & 0.202 & \textbf{0.829} & 0.460 \\
H6  & 0.703 & 0.160 & 0.659 & 0.137 & 0.800 & 0.392 & 0.534 & 0.094 & 0.725 & 0.240 & 0.710 & 0.220 & 0.771 & 0.198 & \textbf{0.833} & 0.467 \\
H7  & 0.699 & 0.160 & 0.653 & 0.136 & 0.784 & 0.336 & 0.607 & 0.122 & 0.723 & 0.239 & 0.721 & 0.220 & 0.768 & 0.196 & \textbf{0.819} & 0.447 \\
H8  & 0.697 & 0.160 & 0.649 & 0.137 & 0.759 & 0.286 & 0.501 & 0.077 & 0.718 & 0.238 & 0.728 & 0.227 & 0.765 & 0.192 & \textbf{0.826} & 0.434 \\
H9  & 0.694 & 0.158 & 0.651 & 0.138 & 0.755 & 0.256 & 0.574 & 0.109 & 0.717 & 0.237 & 0.736 & 0.251 & 0.760 & 0.191 & \textbf{0.824} & 0.426 \\
H10 & 0.692 & 0.152 & 0.666 & 0.150 & 0.722 & 0.221 & 0.626 & 0.126 & 0.719 & 0.240 & 0.739 & 0.249 & 0.764 & 0.201 & \textbf{0.827} & 0.425 \\

\bottomrule
\end{tabular}

\end{table}

\end{landscape}

strategy for producing clinically useful sepsis predictions across varying prediction windows.

\subsection{Federated Learning Results}
Table~\ref{tab:fl_results} summarizes the FL results. FedDyn achieves the best validation loss (0.2480), degrading only 5.8\% relative to the centralized baseline. FedAvg (0.2728, +16.3\%) and FedProx (0.2731, +16.5\%) perform nearly identically, indicating that the static proximal penalty does not help when local drift is limited to a single epoch per round. FedAvgM (0.3578, +52.6\%) underperforms because the momentum mechanism amplifies gradient noise rather than accelerating convergence when the number of local steps per round is small. Note that these FL experiments use a reduced architecture (encoder channels [16, 32], latent dimension 32) for computational efficiency, whereas the main results use the full architecture.

\begin{table*}[t]
\centering
\caption{Federated learning results: best validation loss (BCE) and relative degradation from centralized training. These experiments use a reduced architecture (channels [16, 32], latent dim 32).}
\label{tab:fl_results}
\small
\begin{tabular}{lcccc}
\toprule
\textbf{Method} & \textbf{Val Loss} & \textbf{Abs. $\Delta$} & \textbf{Rel. Deg.} & \textbf{Key Characteristic} \\
\midrule
Centralized (Non-FL) & 0.2345 & --- & --- & Full data distribution \\
FedDyn~\cite{acar2021} & \textbf{0.2480} & +0.0135 & \textbf{+5.8\%} & Per-client dynamic correction \\
FedAvg~\cite{mcmahan2017} & 0.2728 & +0.0383 & +16.3\% & Weighted parameter averaging \\
FedProx~\cite{li2020fedprox} & 0.2731 & +0.0386 & +16.5\% & Static proximal penalty \\
FedAvgM~\cite{hsu2019} & 0.3578 & +0.1233 & +52.6\% & Server momentum \\
\bottomrule
\end{tabular}
\end{table*}

\begin{figure}[t]
\centering
\includegraphics[width=0.9\columnwidth]{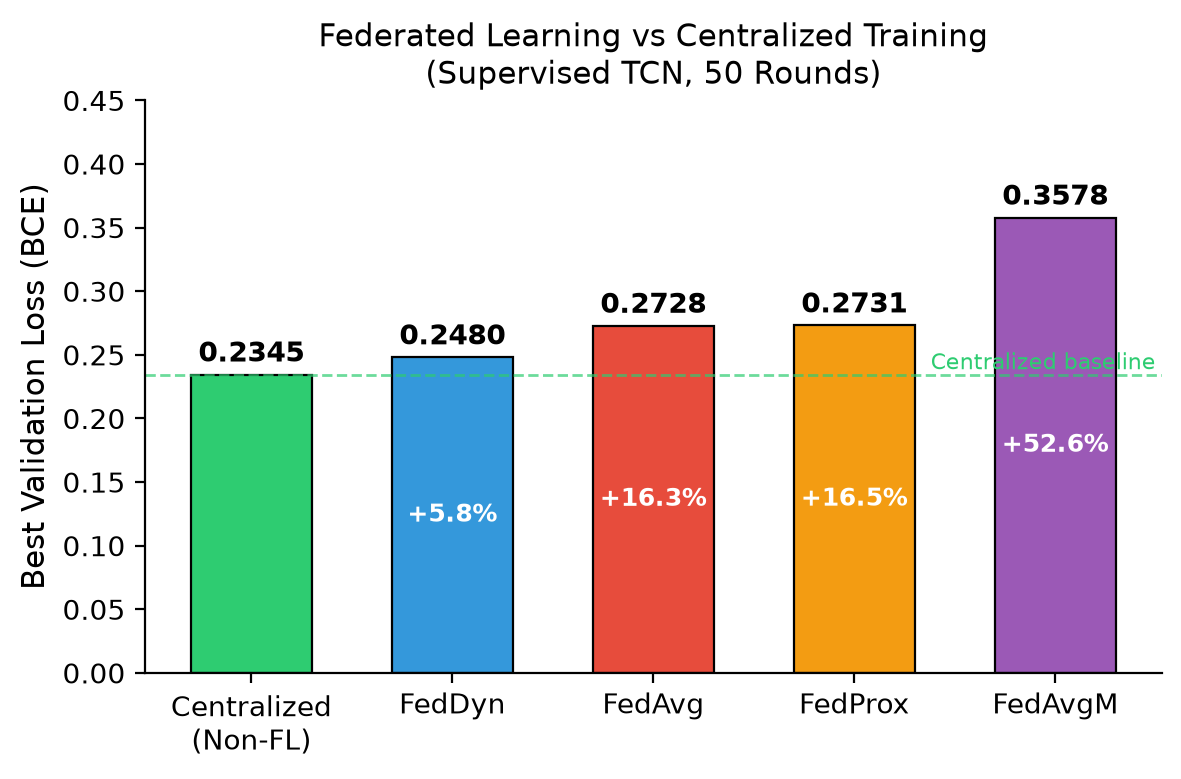}
\caption{Best validation loss (BCE) for centralized training and four federated learning aggregation strategies across 5 clients with LDA($\alpha=0.5$) label distribution. FedDyn closes the gap to within 5.8\% of the centralized baseline.}
\label{fig:fl_comparison}
\end{figure}

\subsection{Comparison with Published Benchmarks}
Table~\ref{tab:comparison} provides a qualitative comparison. Direct numerical comparison is complicated by differences in datasets, biomarker sets, and evaluation protocols.

\begin{table*}[t]
\centering
\caption{Qualitative comparison with published sepsis detection benchmarks. All numbers as reported in respective papers.}
\label{tab:comparison}
\small
\begin{tabular}{lccccc}
\toprule
\textbf{Method} & \textbf{Dataset} & \textbf{Biomarkers} & \textbf{AUROC} & \textbf{AUPRC} & \textbf{Horizon} \\
\midrule
Futoma MGP-RNN~\cite{futoma2017} & Duke EHR & 34 & --- & $\sim$0.50--0.55 & H4 \\
Moor MGP-TCN~\cite{moor2019} & MIMIC-III & 44 & --- & 0.35 & H7 \\
Moor DTW-KNN~\cite{moor2019} & MIMIC-III & 44 & --- & 0.40 & H7 \\
Wang MB-TCN~\cite{wang2022} & PhysioNet 2019 & 40 & 0.892 & 0.527 & Imminent \\
Dai PoEMS~\cite{dai2023} & PhysioNet 2019 & 25 & 0.911 & 0.675 & Imminent \\
Wanyan SupMix~\cite{wanyan2023} & PhysioNet 2019 & 40 & 0.883 & 0.667 AP & H0 \\
\textbf{Ours (JEPA+XGB+mean)} & \textbf{MIMIC-III} & \textbf{7} & \textbf{0.838} & \textbf{0.636} & \textbf{H0} \\
\textbf{Ours (FT+XGB+mean)} & \textbf{MIMIC-III} & \textbf{7} & \textbf{0.850} & \textbf{0.510} & \textbf{H0} \\
\bottomrule
\end{tabular}
\end{table*}

\begin{figure}[t]
\centering
\includegraphics[width=0.9\columnwidth]{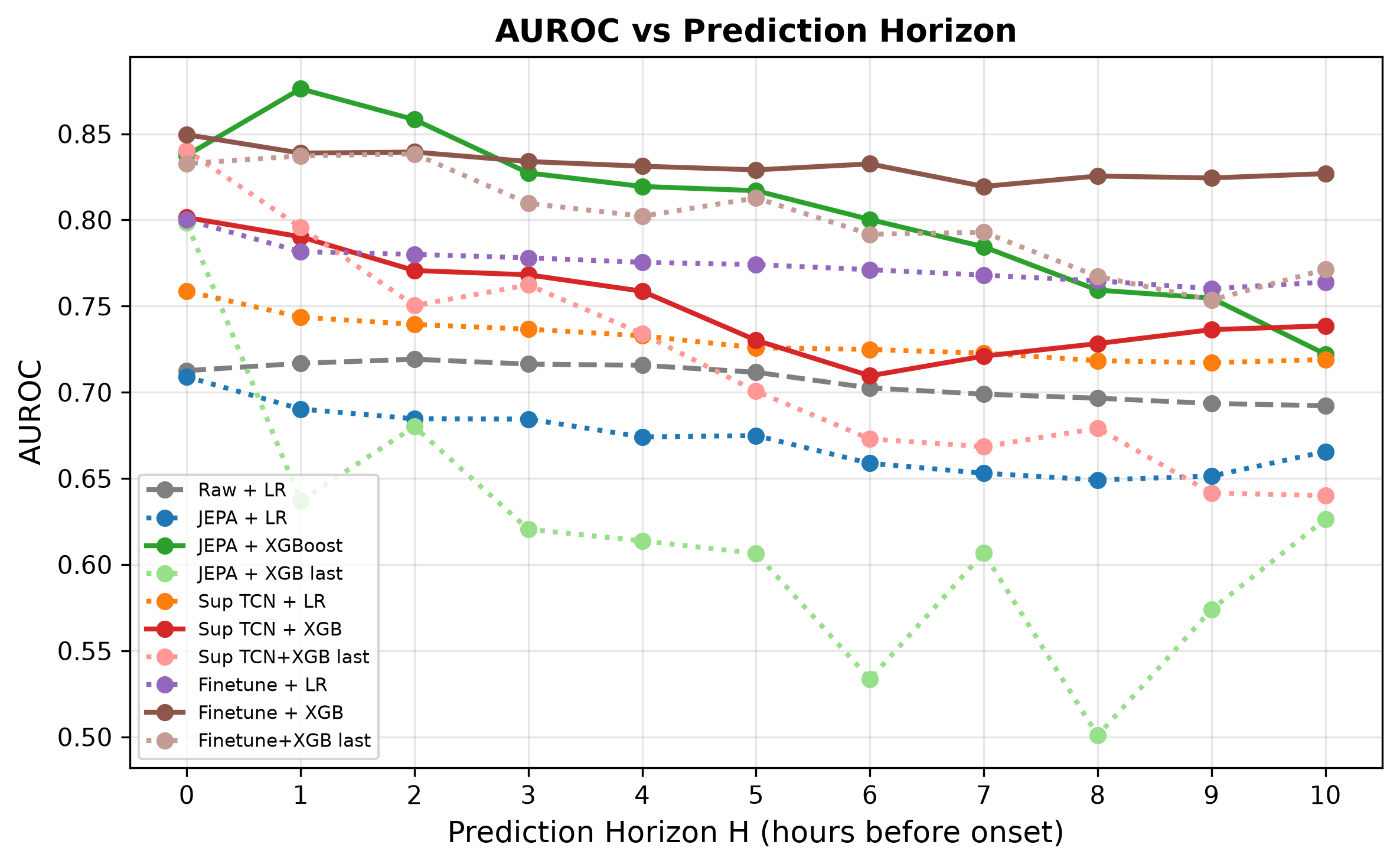}
\caption{AUROC vs Prediction Horizon. $H=0$ (right) = imminent prediction; $H=10$ (left) = 10 hours before onset.}
\label{fig:auroc}
\end{figure}

\begin{figure}[t]
\centering
\includegraphics[width=0.9\columnwidth]{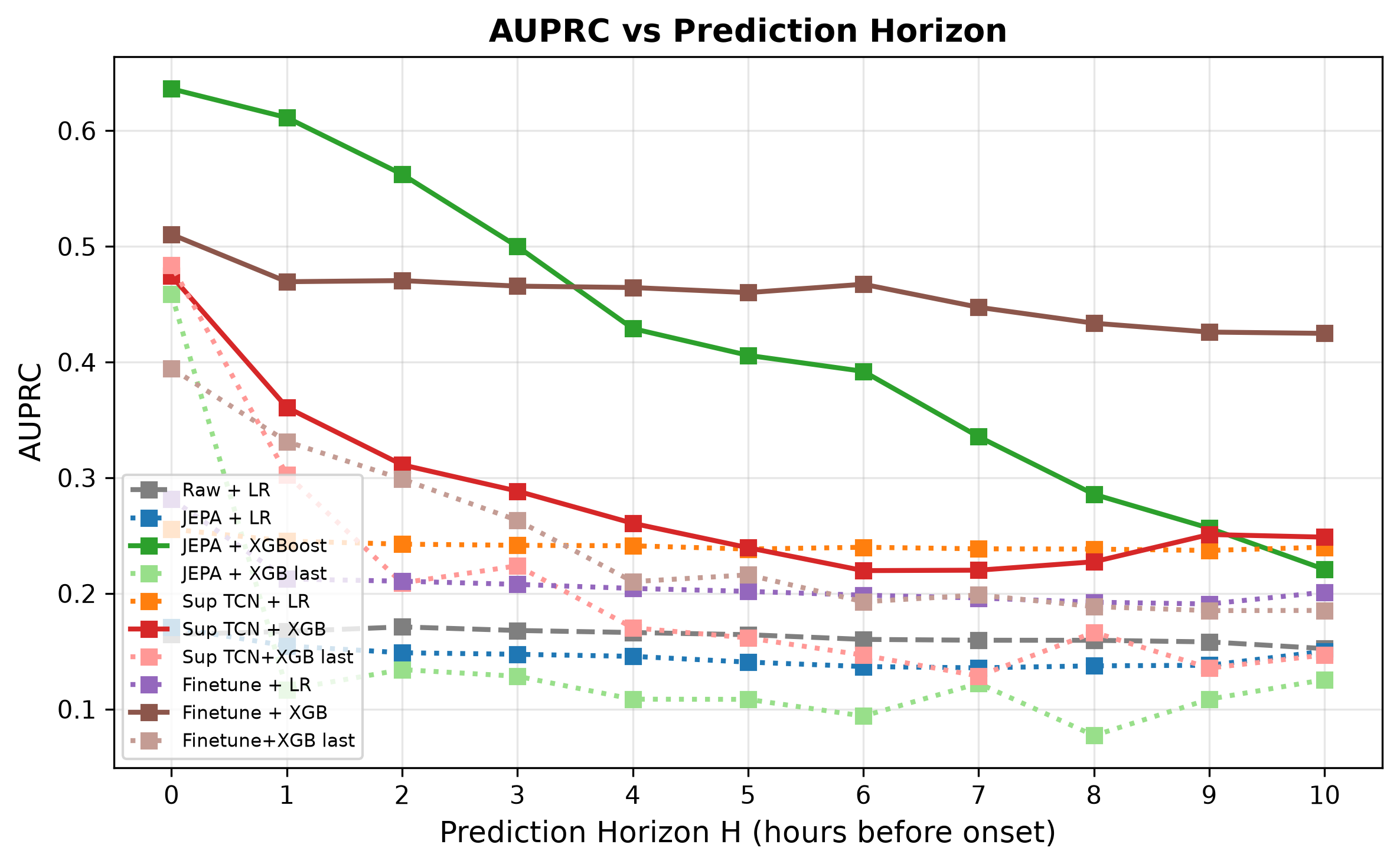}
\caption{AUPRC vs Prediction Horizon across all evaluated models. $H=0$ (right) = imminent prediction; $H=10$ (left) = 10 hours before onset.}
\label{fig:auprc}
\end{figure}

\section{Discussion}

\subsection{Horizon Robustness of Self-Supervised vs.\ Supervised Representations}
The degradation analysis (Table~\ref{tab:degradation}) reveals a clear hierarchy of temporal persistence. The fine-tuned VICReg encoder is the most robust (16.8\% H0$\to$H10 drop), followed by supervised TCN (47.5\%), then JEPA (65.3\%). This ordering directly reflects the degree of task supervision in the training objective.

The JEPA objective---masked latent reconstruction with variance/covariance regularization---excels at modeling local temporal dynamics. The encoder learns to represent the acute physiological changes that immediately precede sepsis onset. At H0, these features are highly discriminative, yielding the overall best AUPRC (0.636). However, the objective does not incentivize the encoder to learn features that are useful at arbitrary distances from the prediction point. When the input window is truncated at H10, the acute signatures are absent and the JEPA features, optimized only for reconstruction fidelity, are poorly suited for classification.

The supervised TCN, by contrast, receives a classification gradient at every time step. The BCE loss penalizes misclassifications based on the pooled representation of the entire window. To minimize this loss, the encoder must develop features that contribute to correct classification regardless of when in the window they occur. This task-aware pressure naturally produces temporally persistent features, reflected in the 47.5\% AUPRC decline.

The fine-tuned VICReg encoder achieves superior robustness (16.8\% drop) because it benefits from both phases. The VICReg pretraining phase learns a rich, general-purpose representation of physiological dynamics through the two-view invariance objective, covering both acute and subtle patterns. The fine-tuning phase then selectively amplifies the discriminative dimensions while preserving the breadth of the pretrained representation. The lower learning rate ($1\times10^{-5}$ vs.\ $5\times10^{-4}$ for supervised TCN) ensures that fine-tuning adjusts rather than overwrites the pretrained features, maintaining the encoder's ability to represent diverse physiological states across the full window.

\subsection{Practical Implications}
\begin{figure}[t]
\centering
\includegraphics[width=0.9\columnwidth]{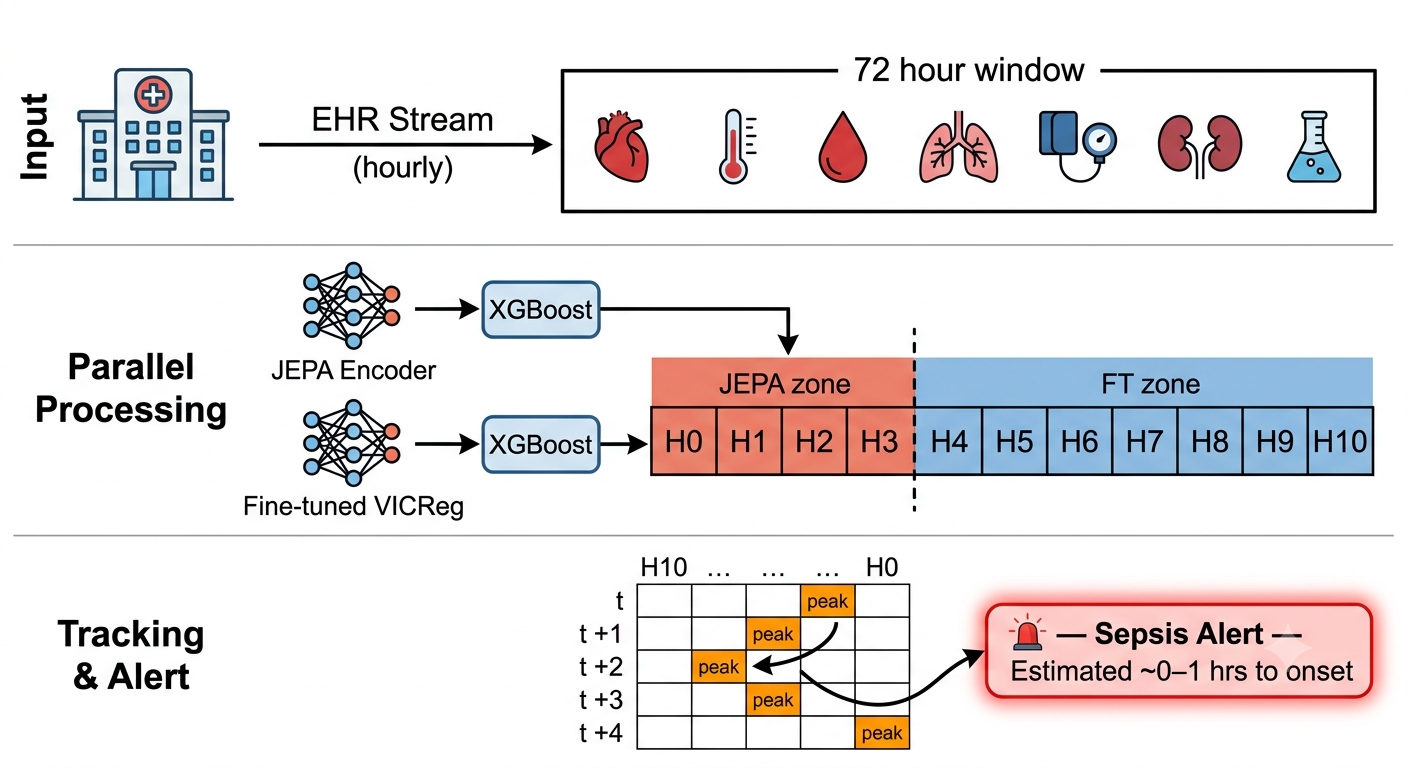}
\caption{Real-time sepsis prediction inference architecture deployed in an ICU setting. EHR data is binned hourly and passed through both the JEPA and fine-tuned VICReg encoders in parallel. Each encoder's embedding is scored by its XGBoost classifier across all 11 horizons simultaneously. A horizon gate selects the appropriate model's output based on the peak horizon (JEPA for H$\leq$3, FT for H$\geq$4). As the patient progresses through their ICU stay, the peak shifts leftward across the horizon profile, providing both an estimated time-to-onset and an escalating risk alert.}
\label{fig:inference}
\end{figure}

The three-tier comparison above identifies a clear performance trade-off across prediction horizons, but translating these results into a clinical deployment requires addressing a fundamental constraint: in an active ICU setting, the prediction horizon H is a \emph{retrospective} label defined relative to a future sepsis onset that has not yet occurred. At any inference time $t$, we do not know how close the patient is to onset, and therefore cannot directly choose which horizon-specific model to apply.

Our deployment architecture, illustrated in Fig.~\ref{fig:inference}, handles this by processing the same hourly-binned 72-hour window through both the JEPA and fine-tuned VICReg encoders in parallel. Each encoder produces a latent representation of the patient's current physiological state, which is then scored by its respective XGBoost classifier at all 11 horizons simultaneously. The result is a full horizon profile $\{s_0, s_1, \dots, s_{10}\}$ where $s_h$ is the risk score for horizon $h$. This profile answers a different question than the horizon-specific evaluation in earlier sections: rather than asking ``what is this model's performance at a known H?'', it asks ``at what temporal distance from a potential onset does this patient's physiology most resemble the training examples for each horizon?'' The horizon with the highest score indicates the most likely temporal position of the patient relative to a future onset, and the magnitude of that peak provides the risk estimate.

A critical design decision is the \emph{horizon gate} that selects which encoder's output to trust at each horizon. Our analysis shows that JEPA+XGB dominates at short horizons (H0--H3, mean AUPRC 0.577) because its masked latent prediction objective captures the acute physiological signatures that immediately precede onset. The fine-tuned encoder dominates at medium-to-long horizons (H4--H10, mean AUPRC 0.446) because the VICReg pretraining preserves broad physiological structure while fine-tuning selects for discriminative patterns that persist earlier in the time series. The gate applies the JEPA profile for H$\leq$3 and the FT profile for H$\geq$4, yielding a hybrid horizon profile that selects the best-performing model at each temporal distance.

The temporal evolution of this profile provides an additional layer of clinical utility. As shown in the tracking table in Fig.~\ref{fig:inference}, the peak horizon shifts leftward over sequential hourly evaluations, tracking the patient's trajectory toward potential onset. A patient whose peak is at H10 at time $t$ but shifts to H7 by $t+3$ is signaling physiological deterioration that warrants increased monitoring, even if absolute risk scores remain below the alert threshold. We propose a three-tier alerting strategy: (i) a \emph{watch} threshold (risk $>$ 0.4) at any H$\geq$4 triggers increased monitoring frequency; (ii) an \emph{alert} threshold (risk $>$ 0.5) at any H$\leq$3 triggers a clinical notification; and (iii) a peak shift of $\geq$2 horizons per hour triggers an escalation review regardless of absolute score. This approach provides clinicians with both a calibrated risk estimate and an estimated time-to-onset window derived from the hybrid horizon profile.

\begin{figure*}[t]
\centering
\includegraphics[width=\textwidth]{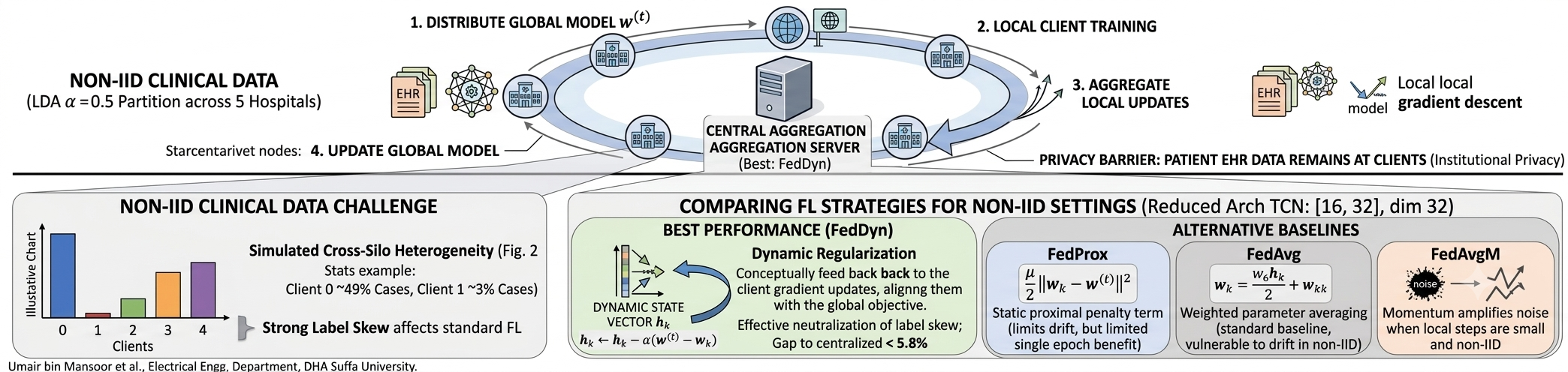}
\caption{Validation loss convergence curves for centralized training and four federated learning aggregation strategies (FedAvg, FedProx, FedAvgM, FedDyn) under LDA($\alpha=0.5$) label distribution across 5 clients. FedDyn consistently tracks the centralized baseline, while FedAvgM diverges due to server momentum amplifying client heterogeneity.}
\label{fig:fl_convergence}
\end{figure*}

\subsection{Federated Learning Under Non-IID Data}
The federated learning results demonstrate that privacy-preserving distributed training can approach centralized performance when the aggregation strategy accounts for client heterogeneity. FedDyn's dynamic per-client correction ($\mathbf{h}_k$ vectors that track the accumulated gradient bias) effectively neutralizes the label skew in our LDA partition, achieving a validation loss of 0.2480---only 5.8\% above the centralized baseline. This is consistent with Acar et al.~\cite{acar2021}, who reported FedDyn gaps of 3--10\% on CIFAR-10 with Dirichlet($\alpha=0.5$) partition. The convergence trajectories of all aggregation strategies are shown in Fig.~\ref{fig:fl_convergence}.

FedProx's static proximal term does not improve over FedAvg (0.2731 vs.\ 0.2728) because with only 1 local epoch per round, per-client drift is minimal. The proximal constraint becomes valuable only when clients take many local steps, which is common in cross-device FL but not in our cross-silo (5-hospital) scenario. FedAvgM underperforms dramatically (0.3578, +52.6\%), as the server momentum mechanism amplifies the natural variance across heterogeneous client updates rather than smoothing it.

Recent work has further advanced FL for sepsis prediction. Chang et al.~\cite{chang2026} integrated FL with a medical knowledge graph and temporal transformer, achieving AUC 0.956 through meta-learning-based personalization across MIMIC-IV and eICU. Fuzail et al.~\cite{fuzail2026} demonstrated that hybrid FL with XGBoost and explainable AI achieves strong predictive performance while preserving data privacy.

These results suggest that for cross-silo clinical FL with small numbers of clients and strong label heterogeneity, FedDyn is the most effective aggregation strategy. The 5.8\% gap to centralized training is a modest price for the privacy and regulatory benefits of keeping patient data within institutional boundaries.

\subsection{Future Work}
Our results demonstrate that a JEPA encoder with forward-fill preprocessing + XGBoost + mean pooling achieves competitive sepsis prediction using only 7 biomarkers, with AUPRC 0.636 approaching the SupMix benchmark (AP 0.667) despite using 83\% fewer markers. The Tier 1 pipeline---VICReg pretraining followed by semi-supervised fine-tuning---achieves AUPRC 0.510 at H0, a 3.1$\times$ improvement over raw features and 7.6\% over supervised TCN, while exhibiting superior horizon robustness (16.8\% H0$\to$H10 drop vs.\ 47.5\% for supervised TCN and 65.3\% for JEPA). The federated learning experiments confirm that FedDyn can close the gap to within 5.8\% of centralized training under realistic label heterogeneity.

Despite these encouraging findings, several promising directions remain for future investigation. First, expanding the biomarker set beyond 7 markers---potentially incorporating the full 34--44 biomarkers used in prior work---could be systematically evaluated with the same SSL pipelines to quantify the saturating returns of additional features. Second, replacing the fixed XGBoost classifier with an end-to-end temporal attention mechanism (e.g., cross-attention between the JEPA/VICReg embeddings and the raw time series) may better exploit the sequential structure that the current mean-pooling step discards. Third, validating the proposed pipelines on multi-center real-world ICU data---rather than simulated FL partitions---would reveal the impact of hardware variability, documentation practices, and population demographics on generalization. Fourth, incorporating uncertainty quantification into the horizon gate (e.g., conformal prediction intervals around each horizon score) could improve the reliability of the peak-horizon selection rule in deployment. Fifth, extending the self-supervised pretraining to a multi-task objective that jointly predicts sepsis onset, organ failure trajectories, and in-hospital mortality may yield representations that are simultaneously informative across related clinical endpoints.

Sixth, adopting VISReg's~\cite{wu2026visreg} sketching-based regularization in place of the covariance term in both JEPA and VICReg objectives may further improve representation quality by enforcing stronger distributional alignment in the latent space.

Seventh, integrating unstructured clinical notes via large language models---as demonstrated by the COMPOSER-LLM system~\cite{shashikumar2025}---could enrich our purely structured-vital-sign representations with contextual information from triage notes and progress reports. Additionally, extending our SSL pipeline to the cross-hospital transfer setting explored by Ding et al.~\cite{ding2023}, or to the trajectory-based deterioration prediction task of Zhang et al.~\cite{zhang2026}, would test the generality of pretrained representations beyond binary sepsis onset classification.

\section{Conclusion}
We presented a comparative methodology for sepsis early prediction comparing JEPA, VICReg with semi-supervised fine-tuning, supervised TCN, and raw feature approaches with LR and XGBoost classifiers. Using 7 carefully selected biomarkers with forward-fill preprocessing, our best model (JEPA + XGBoost + mean pooling) achieves AUPRC 0.636 at H0. The Tier 1 pipeline (VICReg $\to$ fine-tune $\to$ XGBoost) achieves AUPRC 0.510, a 3.1$\times$ improvement over raw features and 7.6\% over supervised TCN. Key findings: (1) VICReg pretraining with semi-supervised fine-tuning produces the most horizon-robust representations (16.8\% H0$\to$H10 drop vs.\ 47.5\% supervised TCN and 65.3\% JEPA); (2) XGBoost consistently outperforms LR; (3) mean pooling beats last-timestep pooling; (4) sparsity-aware biomarker selection (7 markers) achieves results competitive with prior work using 34--44 biomarkers; (5) FedDyn achieves a validation loss of 0.2480 in the federated setting, within 5.8\% of the centralized baseline.

\end{document}